\documentclass{article}

\usepackage[utf8]{inputenc}
\usepackage{microtype}
\usepackage{newunicodechar}
\newunicodechar{ʾ}{'}
\usepackage{graphicx}
\usepackage{subfigure}
\usepackage{booktabs} % for professional tables
\usepackage{multirow}
\usepackage[table,xcdraw]{xcolor}
\usepackage[table]{xcolor}
\usepackage{float}
\usepackage{stfloats}

\usepackage{pifont}
\usepackage{hyperref}

\usepackage[accepted]{icml2025}

\usepackage{amsmath}
\usepackage{amssymb}
\usepackage{mathtools}
\usepackage{amsthm}
\usepackage{enumitem}
\usepackage{pifont}
\usepackage[table]{xcolor}
\usepackage{arydshln}
\usepackage{tikz}
\usepackage[capitalize,noabbrev]{cleveref}
\usepackage{tcolorbox}

\definecolor{LightCyan}{rgb}{0.88,1,1}
\newcommand{\cmark}{\textcolor{green!60!black}{\ding{51}}}
\newcommand{\xmark}{\textcolor{red!70!black}{\ding{55}}}
\newcommand*\circled[1]{\vspace{0.1cm}\tikz[baseline=(char.base)]{
            \node[shape=circle,fill=gray!20,inner sep=1pt] (char) {\textcolor{black}{\small#1}};}}
\newtcolorbox{highlighterbox}[1][]{
    arc=0pt,
    left=2.5pt,
    right=2.5pt,
    bottom=0pt,
    top=0pt, 
    rounded corners,
    boxrule=0.8pt,
    colframe=gray!25,
    colback=gray!5,
}

\theoremstyle{plain}

\theoremstyle{definition}

\theoremstyle{remark}

\usepackage[textsize=tiny]{todonotes}

\icmltitlerunning{From RAG to Agentic: Validating Islamic-Medicine Responses with LLM Agents}

\begin{document}

\twocolumn[
\icmltitle{From RAG to Agentic: Validating Islamic-Medicine Responses with LLM Agents}



\icmlsetsymbol{equal}{*}

\begin{icmlauthorlist}
\icmlauthor{Mohammad Amaan Sayeed}{equal,yyy}
\icmlauthor{Mohammed Talha Alam}{equal,yyy}
\icmlauthor{Raza Imam}{equal,yyy}
\icmlauthor{Shahab Saquib Sohail}{sch}
\icmlauthor{Amir Hussain}{comp}
\end{icmlauthorlist}

\icmlaffiliation{yyy}{Mohamed bin Zayed University of Artificial Intelligence, UAE}
\icmlaffiliation{comp}{Edinburgh Napier University, UK}
\icmlaffiliation{sch}{VIT Bhopal University, India}

\icmlcorrespondingauthor{Mohammed Talha Alam}{mohammed.alam@mbzuai.ac.ae}

\icmlkeywords{Machine Learning, ICML}

\vskip 0.3in
]



\printAffiliationsAndNotice{\icmlEqualContribution} % otherwise use the standard text.

\begin{abstract}
Centuries-old Islamic medical texts like Avicenna’s Canon of Medicine and the Prophetic Tibb-e-Nabawi, encode a wealth of preventive care, nutrition, and holistic therapies, yet remain inaccessible to many and underutilized in modern AI systems. Existing language-model benchmarks focus narrowly on factual recall or user preference, leaving a gap in validating culturally grounded medical guidance at scale. We propose a unified evaluation pipeline, Tibbe-AG, that aligns 30 carefully curated Prophetic-medicine questions with human-verified remedies and compares three LLMs (LLaMA-3, Mistral-7B, Qwen2-7B) under three configurations: direct generation, retrieval-augmented generation, and a \textit{scientific} self-critique filter. Each answer is then assessed by a secondary LLM serving as an agentic judge, yielding a single 3C3H quality score. Retrieval improves factual accuracy by 13\%, while the agentic prompt adds another 10\% improvement through deeper mechanistic insight and safety considerations. Our results demonstrate that blending classical Islamic texts with retrieval and self-evaluation enables reliable, culturally sensitive medical question-answering.	
\end{abstract}
\vspace{-20.5pt}
\section{Introduction}

% How Islam supports health advice, complimenting medical sciences and use of AI in both until now	
Islamic literatures have contributed significantly to medical science. Classic texts written in the 11\textsuperscript{th} century, such as The Canon of Medicine by Avicenna \cite{avicenna2005canon} and Prophetic Tibb-e-Nabawi \cite{saeedgrapes, junaid2019analysis, mohammad1983tibb}, provide detailed guidelines for preventive care, balanced lifestyle, and holistic health. The Canon of Medicine established systematic observation, diagnosis, and treatment protocols that influenced both Eastern and Western practices, while Tibb-e-Nabawi preserves prophetic wisdom by emphasizing hygiene, nutrition, and natural remedies. Historically confined to manuscripts, these sources continue to shape modern healthcare by promoting ethical and patient-centered care. 

\textbf{Research Gap:} LLM-based medical agents, grounded in Tibb-e-Nabawi and The Canon of Medicine, can serve as interactive tools for educating students and scholars of Unani medicine. By integrating these sources into modern language models, it is possible to achieve enhanced contextual understanding that combines clinical data with centuries of ethical and holistic medical wisdom. Such agents can also generate diagnostic recommendations and therapeutic guidance that are both scientifically robust and culturally sensitive, addressing limitations observed in current Unani practices \cite{mehdi2022review}. Ultimately, if scaled effectively, these agents could deliver accessible, personalized, and ethically grounded healthcare solutions that benefit diverse populations from regions like Indian-subcontinent where a huge portion of people rely on Unani medicine \cite{pernau2012indian}, while honoring a rich cultural heritage.

\textbf{Related Work:} 
% Recent AI-based healthcare models have transformed clinical diagnostics and personalized treatment strategies . 
Computational medicine integrated with age-old medical insights can enhance diagnostic precision and therapeutic outcomes \cite{esteva2017dermatologist}.For example, combining traditional knowledge with modern diagnostics improves treatment accuracy \cite{zhang2020artificial}, and deep learning can extract and operationalize data from traditional remedies \cite{zhang2023advances}. Furthermore, \cite{chammas2022deep,alrehali2020historical} integrated Arabic manuscripts points and learning models toward semantic extraction of Islamic texts, bridging the gap between historical wisdom and contemporary computational methods.

\textbf{Contribution:}
Overall, we introduce Tibbe-AG, an agentic RAG framework that blends dense retrieval from classical Islamic medical texts with an explicit self-critique prompt to the same base LLM, yielding scientifically validated, culturally grounded healthcare guidance. We curate a focused benchmark of 30 Prophetic-medicine QA pairs drawn from \emph{Tibb-e-Nabawi} \cite{Shamsi2016} and \cite{5minMadrasa}, and used 3C3H \cite{huggingface3c3h} with secondary LLMs as judge. Experiments across three base LLMs demonstrate that Tibbe-AG significantly outperforms both direct inference and standard RAG baselines.  

\begin{figure*}
    \centering
    \includegraphics[width=0.94\linewidth]{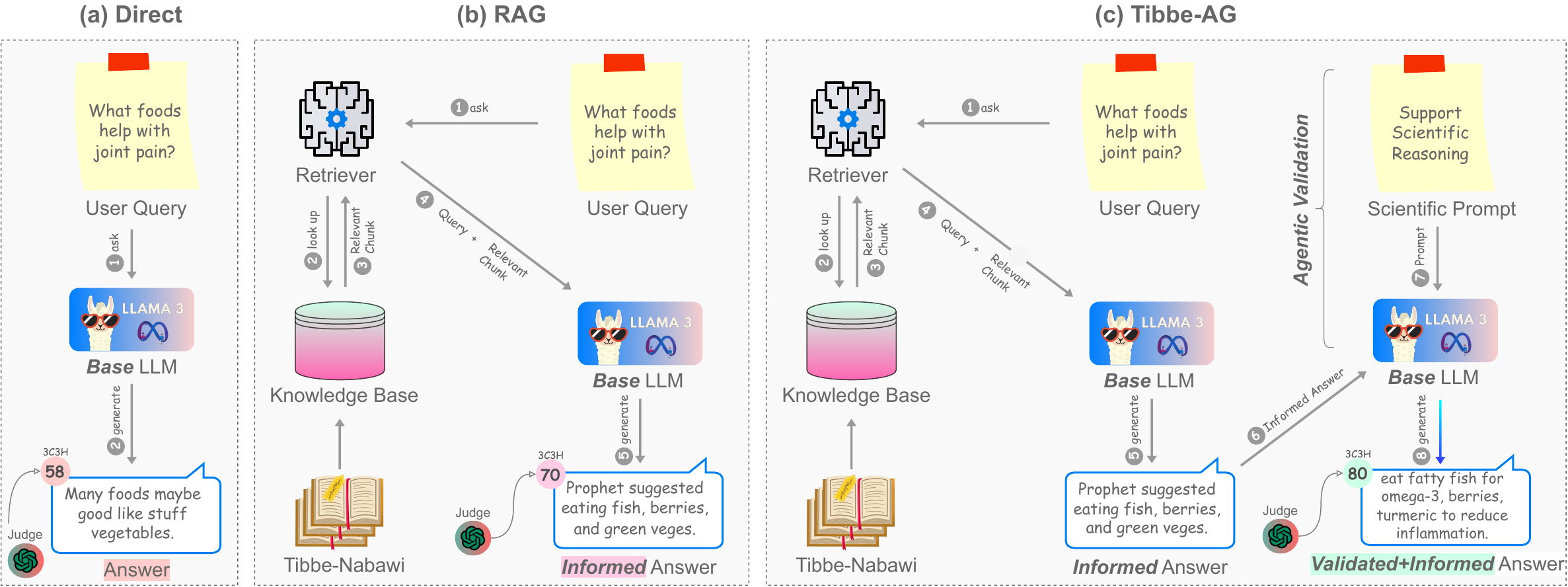}
    \caption{
    \textbf{Overview of TibbeAG:}  (a) In the Direct inference, the base LLM generates an answer from the user query alone, often producing unvalidated or hallucinated content. (b) The RAG setting augments the prompt with top-k passages from the Tibb-e-Nabawi corpus to ground the response, yet still lacks an explicit mechanism to verify factual consistency or safety. (c) TibbeAG combines dense retrieval with an additional self-critique prompt to the same base LLM, yielding a final answer that is scientifically validated.
    }
    \vspace{-0.15cm}
    \label{fig:method}
\end{figure*}

% \section{Method}
% Figures : 1) RAG and Agentic architecture 2) (Optional) 3c3h performance chart for 3 llms
\newpage
\section{Methodology}
\label{sec:method}
We propose a question–answering pipeline, Tibbe-AG, grounded in classical Islamic medical knowledge such as \emph{Tibb-e-Nabawi}. Given a health‐related question \(q\) (for example, \texttt{What foods help with joint pain?}), our framework first retrieves relevant passages from a curated knowledge base and then employs a two‐stage LLM process to generate and validate a final, high‐quality answer.

\textbf{Retrieval:}  
To ensure that generated answers remain firmly rooted in authentic Prophetic‐medicine teachings, we employ a dense retriever i.e., ChromaDB, that converts both query and corpus passages into a shared embedding space via a Transformer‐based encoder. Formally, let
\vspace{-0.2cm}
\begin{equation}
R(q) = \{ d_1, d_2, \dots, d_k \}, 
\quad d_i \in K,
\vspace{-0.2cm}
\end{equation}
denote the top-\(k\) passages selected by cosine‐similarity ranking. Each \(d_i\) is drawn from a preprocessed corpus \(K\) of classical texts, annotated with metadata (e.g., citation, context, severity). By thresholding similarity scores and filtering out redundant or low‐information fragments, this step guarantees that the downstream LLM sees only the most germane and credible evidence, minimizing off‐topic or factual drift.

\textbf{Initial Answer (\(A_0\)):}  
The first \textit{Base} model, \(\mathrm{LLM}_0\), ingests the concatenated query $q$ and retrieved context $R(q)$ to output answer $A_0$ as:
\begin{equation}
A_0 = \mathrm{LLM}_0(q,\,R(q)).
\end{equation}
Under the hood, \(\mathrm{LLM}_0\) is prompted with a structured template that interleaves user question, document excerpts, and explicit `extract-and-summarize' instructions. Internally, it attends over each passage’s provenance tokens to align outputs with source assertions. The resulting draft \(A_0\) synthesizes actionable recommendations (e.g., \texttt{consume 1-2 teaspoons of raw honey daily}) while preserving direct textual traces, thus facilitating traceability and citation of classical sources.

\textbf{Refinement (\(A_f\)):}  
To guard against hallucinations and to enrich mechanistic and safety rationale, we append an explicit validation prompt \(q_{\mathrm{val}}\) to the base LLM’s input. The final answer is then:
\begin{align}
A_f 
&= \mathrm{LLM}_0(q,\,R(q),\,A_0,\,q_{\mathrm{val}})\nonumber\\
&= \mathrm{LLM}_0(q,\,R(q),\,\mathrm{LLM}_0(q,\,R(q)),\,q_{\mathrm{val}}).
\end{align}
This agentic step performs three sub‐tasks using validation prompt as it directs the model to (i) fact-check each \(A_0\) segment against \(R(q)\), (ii) inject mechanistic context (e.g.\ ginger’s effect on COX-2 pathways), and (iii) filter or flag unsafe recommendations (e.g.\ drug–herb interactions). 
% This self-critique step elevates precision and adds missing explanatory links, yielding an evidence-anchored, scientifically coherent response. 
By prompting the same LLM to re‐evaluate its draft against retrieved evidence and explicitly apply mechanistic and safety checks, the additional validation step curbs hallucinations and ensures more reliable, evidence-anchored, and scientifically coherent response.

\textbf{Evaluation (3C3H):}  
To assess answer quality over a test set of \(n\) samples, we compute the aggregated 3C3H score as
\begin{equation}
\scriptsize  
\mathrm{3C3H}
= \frac{1}{6n} \sum_{i=1}^{n} c_{1i}\,\biggl(1 
+ c_{2i}
+ \frac{c_{3i}-1}{4}
+ \frac{h_{1i}-1}{4}
+ \frac{h_{2i}-1}{4}
+ \frac{h_{3i}-1}{4}\biggr),
\end{equation}
where
  \(c_{1i}\) and \(c_{2i}\) are the Correctness and Completeness scores for sample \(i\),  
  \(c_{3i}\) is Conciseness,  
  \(h_{1i}\), \(h_{2i}\), and \(h_{3i}\) are Helpfulness, Harmlessness, and Honesty,  
  each component lies in \([0,1]\) \cite{elFilali2024rethinking}.  
Each criterion is evaluated by the judge model (e.g., GPT4.5 or Gemini) itself, based on the consistency checks for each answer generated by base LLM.

\section{Dataset and Setup}
\subsection{Dataset}
% To evaluate our Tibbe-AG pipeline, we constructed a focused question-answering dataset of \textit{30 Prophetic-medicine questions} drawn from two classical sources:
Validating Islamic-medicine responses via LLMs demands a specialized benchmark that captures the unique terminology and safety considerations of Prophetic treatments. 
To this end, we focused on a carefully curated question-answering dataset of \textit{30 Prophetic-medicine questions}, to assess our experimental setup, drawn from two classical sources: \vspace{-0.2cm}
\begin{highlighterbox}
\small
\begin{itemize}[leftmargin=*, itemsep=0pt, label={}]
    \item\circled{1} \textit{Cures from the Qur’aan and Rasulullaah} as presented in \textit{Madrasah in Just 5 Minutes} \cite{5minMadrasa}, 
    \item\circled{2} \textit{Tibb-e-Nabawi} (Medical Guidance \& Teachings of Prophet Muhammed) \cite{Shamsi2016}.
\end{itemize}
\end{highlighterbox}
Specifically, our curation process proceeded in three steps:
\begin{itemize}[leftmargin=*, itemsep=0pt, label={},align=left, itemsep=0pt, parsep=0pt, topsep=-10pt]
  \item\circled{A} \textbf{Section Extraction:}  
    From the first source, we extracted the entire section titled \textit{“(9) Cures from the Qur’aan and Rasulullaah”} using a PyMuPDF-based script, isolating all remedy descriptions and their original citations. 
    From the second source, we parsed the Prophetic-medicine chapters to compile an analogous list of remedies, ensuring comparable structure (remedy statement + reference).
  \item\circled{B} \textbf{Question Generation and Selection:}  
    We converted each remedy description into a question of the form  ``\texttt{What Prophetic remedy is recommended for <ailment>?}", yielding an initial pool of approximately 120 candidate questions. We then \emph{manually filtered} this pool to 30 questions, balancing across five broad categories, \emph{nutritional therapies, herbal remedies, ritual supplications, hygiene practices,} and \emph{wound treatments}, to ensure coverage of the full spectrum of Islamic-medicine teachings.
\item\circled{C} \textbf{Representativeness \& Feasibility:}  
  Our 30 questions draw equally from Qurʾānic/Rasulullaah-based cures and the broader Tibb-e-Nabawi corpus, thereby spanning both spiritual and herbal dimensions, yet remain tractable for exhaustive evaluation (3 settings × 3 models × 3 judges). Each prompt carries its exact source (e.g. Sūrah, ḥadīth collection, or Tibb-e-Nabawi chapter and verse), enabling precise lookup and citation during retrieval and self-critique.
\end{itemize}

\subsection{Evaluation Setup}

To empirically validate the superiority of our agentic Tibbe‑AG framework, we compare \textit{three inference settings} on the same test questions $q$ from our dataset as:
\begin{enumerate}[label={}, leftmargin=0pt, labelwidth=*, align=left, itemsep=0pt, parsep=0pt, topsep=0pt]
    \item\circled{E1} In the \textbf{Direct} setting\footnote{We use `\textit{Direct}' throughout to denote direct inference, where model receives only the user query and directly outputs an answer.}, the base model receives only the question \(q\) and produces an answer \(A_f = \mathrm{LLM}_0(q)\) without any external grounding.
    \item\circled{E2} The \textbf{RAG} setting augments the prompt with top‑\(k\) passages retrieved from the Tibb‑e‑Nabawi corpus, yielding\\ \(A_f = \mathrm{LLM}_0(q,\,R(q))\).
    \item\circled{E3} Finally, our \textbf{Tibbe‑AG} (Agentic) pipeline applies the same retrieval step as {E2} and but additionally prompts $A_0$ with $q_\mathrm{val}$ to refine the draft answer as: \\ \(A_f = \mathrm{LLM}_0(q,\,R(q),\,A_0,\,q_{\mathrm{val}})\). Detailed in Sec. \ref{sec:method}.
\end{enumerate}
  
This controlled comparison among three settings is essential: by isolating the effects of plain inference, retrieval grounding, and agentic self‑critique, we can precisely quantify each component’s contribution under the 3C3H metric. Only through this three‑way ablation can we demonstrate that retrieval alone improves factual grounding, and that the subsequent judge step yields statistically significant gains in Completeness, Harmlessness, and overall answer reliability. 
% Beyond quantitative validation, this setup also highlights Tibbe‑AG’s unique advantages in enforcing mechanistic explanations and safety checks, critical for culturally sensitive medical advice.

% \vspace{0.5em}
\begin{table}[t]
\centering
\small
\renewcommand{\arraystretch}{1.3}
% \rowcolors{2}{gray!10}{white}
\caption{
    Response quality across Direct, RAG, and Tibbe-AG on four key criteria. Only Tibbe-AG meets all, demonstrating both evidence grounding and safety.
    }
    \vspace{0.1cm}
\begin{tabular}{l | c c >{\columncolor{gray!10}}c}
\hline
% \rowcolor{gray!20}
Response Generation $\rightarrow$ & \textbf{Direct} & \textbf{RAG} & \textbf{Tibbe‑AG} \\
\hline
Cites authentic sources        & \xmark & \cmark & \cmark \\
Provides actionable specifics  & \xmark & \cmark & \cmark \\
Includes scientific validation  & \xmark & \xmark & \cmark \\
Includes clinical safety cues  & \xmark & \xmark & \cmark \\
\hline
\end{tabular}
\label{tab:qa_quality_matrix}
\vspace{-0.4cm}
\end{table}

\subsection{Base and Judge Models}
For the base‐model component of each pipeline, we conduct experiments using three state‐of‐the‐art 7B‑parameter LLMs: \textbf{Mistral‑7B} \cite{jiang2023mistral7b},\textbf{Qwen‑7B} \cite{bai2023qwentechnicalreport}, and \textbf{LLaMA‑3‑7B} \cite{grattafiori2024llama3herdmodels}. In our primary evaluation, the judge model is instantiated as GPT4.5 \cite{openai2024gpt4technicalreport}, which critiques and refines every answer generated by the Direct, RAG, and Tibbe‑AG settings. To verify that our findings are not idiosyncratic to a single judge, we also perform ablation studies replacing \textbf{GPT‑4.5} with 3 alternative evaluators, \textbf{O4-mini-hi}, \textbf{Claude-4} \cite{claude3modelcard} and \textbf{Gemini} \cite{geminiteam2025geminifamilyhighlycapable}, while keeping all other components fixed. This multi‑judge protocol demonstrates that Tibbe‑AG’s improvements in 3C3H scores remain consistent across different evaluation models, reinforcing the robustness of our agentic framework.
% compared to Direct inference or plain RAG. 

\begin{figure}[h]
\centering
\includegraphics[width=0.8\linewidth]{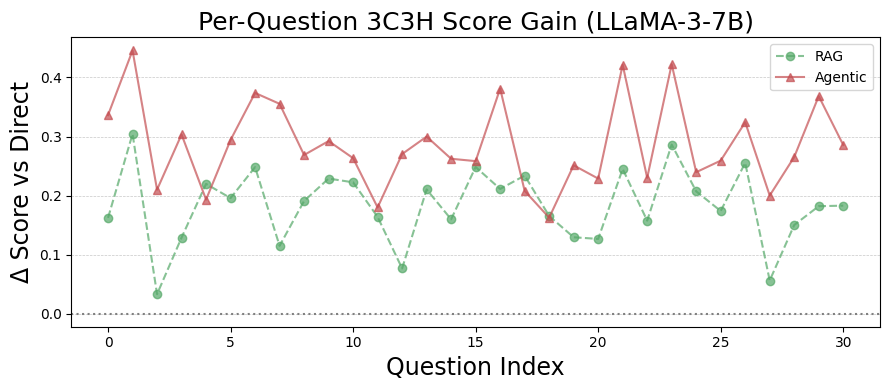}
\vspace{-0.45cm}
\caption{Comparison of \textit{per-sample 3C3H score} gain among: Direct, RAG, and Agentic. Each point represents a unique query, highlighting performance improvements with \textit{Tibbe-AG} (Agentic).}
\label{fig:3c3h_gain}
\end{figure}

% \begin{figure}[t]
% \centering
% \includegraphics[width=0.9\linewidth]{Figures/avg-3c3h.png}
% \caption{Average 3C3H scores (Correctness, Coherence, Coverage, Helpfulness, Harm Avoidance) across the three settings. Tibbe‑AG demonstrates consistent improvements in both faith-based and scientific dimensions.}
% \label{fig:3c3h_avg}
% \end{figure}

% \usepackage{multirow}
% {\renewcommand{\arraystretch}{1.3}
% \begin{table}[]
% \caption{Average 3C3H scores (Correctness, Coherence, Coverage, Helpfulness, Harm Avoidance) across the three settings. Tibbe‑AG demonstrates consistent improvements in both faith-based and scientific dimensions.}
% \label{tab:main}
% \resizebox{0.50\textwidth}{!}{%
% \begin{tabular}{ll|llll}
% \hline
% {Base Models~$\downarrow$} & {Methods~$\downarrow$} & \textbf{gpt4.5} & \textbf{o4-mini-hi} & \textbf{claude} & \textbf{mean} \\ \hline
% \multirow{3}{*}{\textbf{Qwen}} & Direct & 0.44 &  0.48 &  &  \\
%  & RAG & \underline{0.62} & \underline{0.62} &  &  \\
%  & \rowcolor{gray!10}{Tibbe-AG} & \textbf{0.73} & \textbf{0.74} &  &  \\ \hline
% \multirow{3}{*}{\textbf{Mistral}} & Direct & 0.48 &  0.53 &  &  \\
%  & RAG & 0.65 & \underline{0.66} &  &  \\
%  & \rowcolor{gray!10}{Tibbe-AG} & \textbf{0.76} & \textbf{0.77} &  &  \\ \hline
% \multirow{3}{*}{\textbf{LLaMA-3}} & Direct & 0.49 & 0.58  &  &  \\
%  & RAG & 0.67 & \underline{0.70} &  &  \\
%  & \rowcolor{gray!10}{Tibbe-AG} & \textbf{0.77} & \textbf{0.79} &  & \\ \hline
% \end{tabular}}
% \end{table}}

\begin{figure*}[t]
    \centering
    \includegraphics[width=0.91\linewidth]{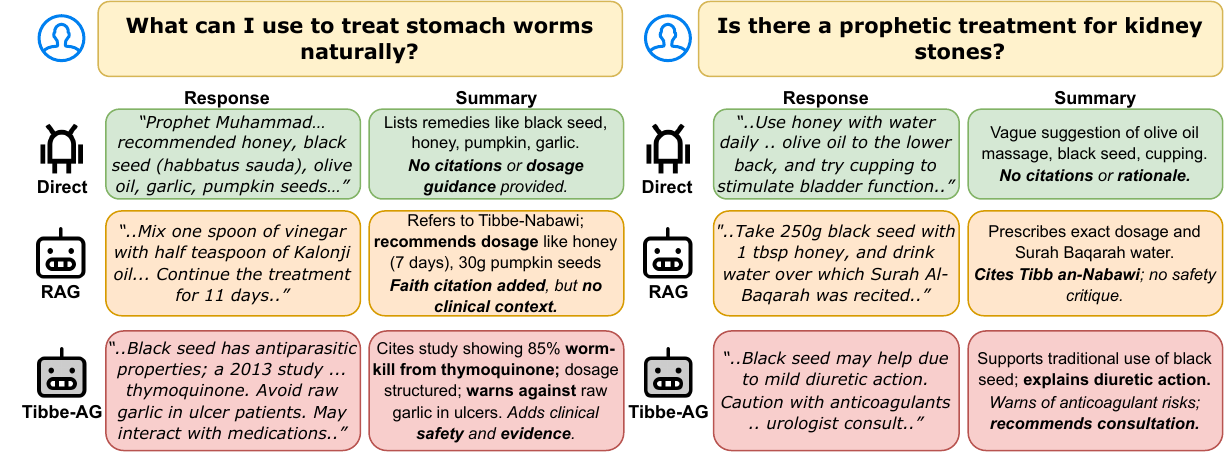}
    \vspace{-0.2cm}
    \caption{
    \textbf{Qualitative comparison of representative response excerpts across three inference settings.} Direct responses list remedies without justification; RAG adds faith-based specificity and references; Tibbe-AG integrates scientific critique, safety reasoning, and actionable guidance.
    }
    \label{fig:vis-tibbe-ag}
    \vspace{-0.4cm}
\end{figure*}

\renewcommand{\arraystretch}{1.2}
\begin{table}[t]
  \vspace{-0.15cm}
  \caption{Average 3C3H scores across inference settings. Tibbe-AG demonstrates consistent improvements in both faith-based and scientific dimensions. \textbf{Mean} is the average of four Judge LLMs.}
  \vspace{0.2cm}
  \label{tab:main}
  \centering
  \resizebox{0.48\textwidth}{!}{%
    \begin{tabular}{l|l|cccc|c}
      \hline
      \textbf{Base Model~$\downarrow$} 
        & \textbf{Method~$\downarrow$} 
        & \textbf{GPT4.5} 
        & \textbf{O4-mini} 
        & \textbf{Claude-4} 
        & \textbf{Gemini} 
        & \textbf{Mean} 
      \\ \hline
      \multirow{3}{*}{Qwen} 
        & Direct     & 0.44 & 0.48 & 0.43 & 0.44 & 0.45 \\ 
        & RAG        & \underline{0.62} & \underline{0.62} & \underline{0.75} & \underline{0.76} & \underline{0.69} \\ 
      \rowcolor{gray!10}
        \cellcolor{white} & Tibbe-AG   & \textbf{0.73} & \textbf{0.74} & \textbf{0.85} & \textbf{0.86} & \textbf{0.80} \\ \hline
      \multirow{3}{*}{Mistral} 
        & Direct     & 0.48 & 0.53 & 0.45 & 0.46 & 0.48 \\ 
        & RAG        & 0.65 & \underline{0.66} & \underline{0.76} & \underline{0.77} & \underline{0.71} \\ 
      \rowcolor{gray!10}
        \cellcolor{white} & Tibbe-AG   & \textbf{0.76} & \textbf{0.77} & \textbf{0.86} & \textbf{0.87} & \textbf{0.82} \\ \hline
      \multirow{3}{*}{LLaMA-3} 
        & Direct     & 0.49 & 0.58 & 0.47 & 0.48 & 0.50 \\ 
        & RAG        & 0.67 & \underline{0.70} & \underline{0.78} & \underline{0.79} & \underline{0.73} \\ 
      \rowcolor{gray!10}
        \cellcolor{white} & Tibbe-AG   & \textbf{0.77} & \textbf{0.79} & \textbf{0.88} & \textbf{0.89} & \textbf{0.83} \\ \hline
    \end{tabular}%
  }% end \resizebox
  \vspace{-0.5cm}
\end{table}

\section{Results and Discussion}

\textbf{Quantitative Results:}
As shown in Table \ref{tab:main}, Tibbe-AG consistently achieved higher average 3C3H scores across all tested base models (LLaMA-3, Mistral-7B, Qwen2-7B) when compared to both Direct inference and standard RAG approaches.  For instance, with LLaMA-3 as the base model and GPT4.5 as the judge, Tibbe-AG scored 0.77, surpassing Direct (0.49) and RAG (0.67) by significant margins.  This trend of improvement is robust across different judge models, as the mean results are consistently higher with our approach, indicating that the benefits of Tibbe-AG are not tied to a specific evaluator, reinforcing generalizability.

\textbf{Qualitative Results:}
Direct responses often lack grounding and specifics, while RAG, though an improvement, typically falls short on scientific validation and safety considerations. Figure \ref{fig:vis-tibbe-ag} provides examples of these distinctions.  For instance, when asked about treating stomach worms, the Direct response is generic.  The RAG response offers more specific remedies from Tibb-e-Nabawi but lacks clinical context or safety warnings.  Tibbe-AG, however, not only suggests remedies like black seed but also cites a relevant study on thymoquinone's efficacy, provides dosage considerations, and crucially, warns about contraindications (e.g., raw garlic in ulcer patients) and potential medication interactions.  Similarly, for kidney stones, Tibbe-AG explains the potential mechanism (diuretic action of black seed), warns about anticoagulant risks, and advises consulting a urologist, demonstrating a more comprehensive and responsible approach.

\textbf{Discussion:}
The enhanced performance of Tibbe-AG can be attributed to its two-stage process. First, the retrieval step grounds the LLM in the relevant classical Islamic medical texts, which helps keep the information accurate and true to the source. This directly combats the tendency for LLMs to "hallucinate" when generating answers directly. Then comes the crucial agentic self-critique. By prompting the LLM to review its initial answer against the retrieved evidence and to actively think about mechanistic details and safety, Tibbe-AG polishes the response into something more coherent, scientifically plausible, and safe. This design directly addresses a key shortcoming in many existing systems, which often don't adequately validate medical guidance that's grounded in specific cultural contexts.

\section{Conclusion}
Our work successfully introduces Tibbe-AG, a novel framework that significantly enhances the generation of reliable and culturally sensitive Islamic medical guidance using LLMs. By effectively blending classical texts with retrieval and agentic self-critique, Tibbe-AG paves the way for AI systems that make the rich heritage of Islamic medicine more accessible and applicable in modern contexts. While our initial 30-question dataset  provided a strong foundation, future efforts will focus on expanding this dataset, further refining Tibbe-AG’s agentic capabilities, and conducting user studies. These steps will be crucial in evolving Tibbe-AG into an even more robust tool, contributing to a more inclusive and culturally competent approach in healthcare.

\newpage
\bibliography{example_paper}

\begin{thebibliography}{21}
\providecommand{\natexlab}[1]{#1}
\providecommand{\url}[1]{\texttt{#1}}
\expandafter\ifx\csname urlstyle\endcsname\relax
  \providecommand{\doi}[1]{doi: #1}\else
  \providecommand{\doi}{doi: \begingroup \urlstyle{rm}\Url}\fi

\bibitem[Alrehali et~al.(2020)Alrehali, Alsaedi, Alahmadi, and Abid]{alrehali2020historical}
Alrehali, B., Alsaedi, N., Alahmadi, H., and Abid, N.
\newblock Historical arabic manuscripts text recognition using convolutional neural network.
\newblock In \emph{2020 6th conference on data science and machine learning applications (CDMA)}, pp.\  37--42. IEEE, 2020.

\bibitem[Anthropic(2024)]{claude3modelcard}
Anthropic.
\newblock The claude 3 model family: Opus, sonnet, haiku, 2024.
\newblock URL \url{https://www-cdn.anthropic.com/de8ba9b01c9ab7cbabf5c33b80b7bbc618857627/Model_Card_Claude_3.pdf}.
\newblock Accessed: May 20, 2025.

\bibitem[Avicenna(2005)]{avicenna2005canon}
Avicenna.
\newblock \emph{The Canon of Medicine}.
\newblock Kazi Publications, 2005.

\bibitem[Bai et~al.(2023)Bai, Bai, Chu, Cui, et~al.]{bai2023qwentechnicalreport}
Bai, J., Bai, S., Chu, Y., Cui, Z., et~al.
\newblock Qwen technical report, 2023.
\newblock URL \url{https://arxiv.org/abs/2309.16609}.

\bibitem[Chammas et~al.(2022)Chammas, Makhoul, Demerjian, and Dannaoui]{chammas2022deep}
Chammas, M., Makhoul, A., Demerjian, J., and Dannaoui, E.
\newblock A deep learning based system for writer identification in handwritten arabic historical manuscripts.
\newblock \emph{Multimedia Tools and Applications}, 81\penalty0 (21):\penalty0 30769--30784, 2022.

\bibitem[El~Filali et~al.(2024)El~Filali, Sengupta, Abouelseoud, Nakov, Fourrier, AI, and MBZUAI]{elFilali2024rethinking}
El~Filali, A., Sengupta, N., Abouelseoud, A., Nakov, P., Fourrier, C., AI, I., and MBZUAI.
\newblock Rethinking {LLM} evaluation with 3c3h: Aragen benchmark and leaderboard, December 2024.
\newblock URL \url{https://huggingface.co/blog/leaderboard-3c3h-aragen}.
\newblock Accessed: 2025-05-30.

\bibitem[Esteva et~al.(2017)Esteva, Kuprel, Novoa, Ko, Swetter, Blau, and Thrun]{esteva2017dermatologist}
Esteva, A., Kuprel, B., Novoa, R., Ko, J., Swetter, S., Blau, H., and Thrun, S.
\newblock Dermatologist-level classification of skin cancer with deep neural networks.
\newblock \emph{Nature}, 542\penalty0 (7639):\penalty0 115--118, 2017.

\bibitem[Google(2025)]{geminiteam2025geminifamilyhighlycapable}
Google.
\newblock Gemini: A family of highly capable multimodal models, 2025.
\newblock URL \url{https://arxiv.org/abs/2312.11805}.

\bibitem[Grattafiori et~al.(2024)Grattafiori, Dubey, Jauhri, et~al.]{grattafiori2024llama3herdmodels}
Grattafiori, A., Dubey, A., Jauhri, A., et~al.
\newblock The llama 3 herd of models, 2024.
\newblock URL \url{https://arxiv.org/abs/2407.21783}.

\bibitem[{Hugging Face}(2024)]{huggingface3c3h}
{Hugging Face}.
\newblock 3c3h: Arabic reasoning and faith-aligned evaluation leaderboard, 2024.
\newblock URL \url{https://huggingface.co/blog/leaderboard-3c3h-aragen}.
\newblock Accessed: May 14, 2025.

\bibitem[Jiang et~al.(2023)Jiang, Sablayrolles, Mensch, et~al.]{jiang2023mistral7b}
Jiang, A.~Q., Sablayrolles, A., Mensch, A., et~al.
\newblock Mistral 7b, 2023.
\newblock URL \url{https://arxiv.org/abs/2310.06825}.

\bibitem[Junaid \& Ali(2019)Junaid and Ali]{junaid2019analysis}
Junaid, M. and Ali, S.
\newblock Analysis of historical islamic medical manuscripts.
\newblock In \emph{Proceedings of the International Conference on Islamic Sciences}, pp.\  145--150, 2019.

\bibitem[Mehdi et~al.(2022)Mehdi, Sultana, Heyat, Chola, Akhtar, Gutema, Al-qadasi, and Baig]{mehdi2022review}
Mehdi, S., Sultana, A., Heyat, M. B.~B., Chola, C., Akhtar, F., Gutema, H.~K., Al-qadasi, D.~M., and Baig, A.~A.
\newblock A review of amenorrhea toward unani to modern system with emerging technology: current advancements, research gap, and future direction.
\newblock \emph{Computational Intelligence in Healthcare Applications}, pp.\  121--135, 2022.

\bibitem[Mohammad(1983)]{mohammad1983tibb}
Mohammad, A.
\newblock Tibb-e-nabawi: Medical teachings in prophetic tradition.
\newblock \emph{Islamic Journal of Medical History}, 1:\penalty0 34--42, 1983.

\bibitem[{Mufti A.H.Elias}(2012)]{5minMadrasa}
{Mufti A.H.Elias}.
\newblock \emph{5-Minute Madrasa in English}.
\newblock 2012.
\newblock URL \url{https://www.islamicstudiesresources.com/uploads/1/9/8/1/19819855/5-minute-madrasa-in-english.pdf}.
\newblock Accessed: 2025-05-14.

\bibitem[OpenAI(2024)]{openai2024gpt4technicalreport}
OpenAI.
\newblock Gpt-4 technical report, 2024.
\newblock URL \url{https://arxiv.org/abs/2303.08774}.

\bibitem[Pernau(2012)]{pernau2012indian}
Pernau, M.
\newblock The indian body and unani medicine: Body history as entangled history.
\newblock In \emph{Images of the Body in India}, pp.\  97--108. Routledge India, 2012.

\bibitem[Saeed \& Grapes(2015)Saeed and Grapes]{saeedgrapes}
Saeed, A. and Grapes, M.
\newblock Prophetic guidance on health: An analysis of tibb-e-nabawi.
\newblock \emph{Journal of Islamic Medical Studies}, 4\penalty0 (2):\penalty0 98--105, 2015.

\bibitem[Shamsi(2016)]{Shamsi2016}
Shamsi, D. M.~S.
\newblock \emph{Tibb-e-Nabawi: Medical Guidance \& Teachings of Prophet Muḥammad}.
\newblock 2016.
\newblock URL \url{https://ia600507.us.archive.org/29/items/TibENabi1/Tib-e-nabi-%20(1).pdf}.
\newblock Accessed: May 14, 2025.

\bibitem[Zhang et~al.(2020)Zhang, Ni, Li, Zhang, et~al.]{zhang2020artificial}
Zhang, H., Ni, W., Li, J., Zhang, J., et~al.
\newblock Artificial intelligence--based traditional chinese medicine assistive diagnostic system: validation study.
\newblock \emph{JMIR medical informatics}, 8\penalty0 (6):\penalty0 e17608, 2020.

\bibitem[Zhang et~al.(2023)Zhang, Wang, Pi, He, and Liu]{zhang2023advances}
Zhang, S., Wang, W., Pi, X., He, Z., and Liu, H.
\newblock Advances in the application of traditional chinese medicine using artificial intelligence: a review.
\newblock \emph{The American journal of Chinese medicine}, 51\penalty0 (05):\penalty0 1067--1083, 2023.

\end{thebibliography}
\bibliographystyle{icml2025}

%%%%%%%%%%%%%%%%%%%%%%%%%%%%%%%%%%%%%%%%%%%%%%%%%%%%%%%%%%%%%%%%%%%%%%%%%%%%%%%
%%%%%%%%%%%%%%%%%%%%%%%%%%%%%%%%%%%%%%%%%%%%%%%%%%%%%%%%%%%%%%%%%%%%%%%%%%%%%%%
% APPENDIX
%%%%%%%%%%%%%%%%%%%%%%%%%%%%%%%%%%%%%%%%%%%%%%%%%%%%%%%%%%%%%%%%%%%%%%%%%%%%%%%
%%%%%%%%%%%%%%%%%%%%%%%%%%%%%%%%%%%%%%%%%%%%%%%%%%%%%%%%%%%%%%%%%%%%%%%%%%%%%%%
% \newpage
% \appendix
% \onecolumn

% You can have as much text here as you want. The main body must be at most $8$ pages long.
% For the final version, one more page can be added.
% If you want, you can use an appendix like this one.  

% The $\mathtt{\backslash onecolumn}$ command above can be kept in place if you prefer a one-column appendix, or can be removed if you prefer a two-column appendix.  Apart from this possible change, the style (font size, spacing, margins, page numbering, etc.) should be kept the same as the main body.
%%%%%%%%%%%%%%%%%%%%%%%%%%%%%%%%%%%%%%%%%%%%%%%%%%%%%%%%%%%%%%%%%%%%%%%%%%%%%%%
%%%%%%%%%%%%%%%%%%%%%%%%%%%%%%%%%%%%%%%%%%%%%%%%%%%%%%%%%%%%%%%%%%%%%%%%%%%%%%%

\end{document}